\documentclass[sigconf]{acmart}

\usepackage{amssymb}
\usepackage{epsfig}
\usepackage{graphicx}
\usepackage{amsmath}
\usepackage{booktabs}
\usepackage{array}
\usepackage{multirow}
\usepackage{CJK}
\usepackage{float}
\usepackage{balance}
\usepackage{longtable}
\usepackage{pdfpages}
\newcommand{\tabincell}[2]{\begin{tabular}{@{}#1@{}}#2\end{tabular}}

\AtBeginDocument{%
  \providecommand\BibTeX{{%
    \normalfont B\kern-0.5em{\scshape i\kern-0.25em b}\kern-0.8em\TeX}}}

\copyrightyear{2021}
\acmYear{2021}
\setcopyright{acmcopyright}
\acmConference[MM '21] {Proceedings of the 29th ACM International Conference on Multimedia}{October 20--24, 2021}{Virtual Event, China.}
\acmBooktitle{Proceedings of the 29th ACM Int'l Conference on Multimedia (MM '21), Oct. 20--24, 2021, Virtual Event, China}
\acmPrice{15.00}
\acmISBN{978-1-4503-8651-7/21/10}
\acmDOI{10.1145/3474085.3475225}

\begin{document}
\fancyhead{}
\title{ZiGAN: Fine-grained Chinese Calligraphy Font Generation \\ via a Few-shot Style Transfer Approach}

\author{Qi Wen}
\email{wenqijay@gmail.com}
\authornote{Authors contribute equally.}
\affiliation{%
  \institution{NetEase Fuxi AI Lab}
  \city{Hangzhou}
  \country{China}
}

\author{Shuang Li}
\email{shuangli@bit.edu.cn}
\authornotemark[1]
\affiliation{%
  \institution{Beijing Institute of Technology}
  \city{Beijing}
  \country{China}
}

\author{Bingfeng Han}
\email{bfhan@bit.edu.cn}
\affiliation{%
  \institution{Beijing Institute of Technology}
  \city{Beijing}
  \country{China}
}

\author{Yi Yuan}
\email{yuanyi@corp.netease.com}
\authornote{Dr. Yuan is the corresponding author.}
\affiliation{%
  \institution{NetEase Fuxi AI Lab}
  \city{Hangzhou}
  \country{China}
}

\begin{abstract}
   Chinese character style transfer is a very challenging problem because of the complexity of the glyph shapes or underlying structures and large numbers of existed characters, when comparing with English letters. Moreover, the handwriting of calligraphy masters has a more irregular stroke and is difficult to obtain in real-world scenarios. Recently, several GAN-based methods have been proposed for font synthesis, but some of them  require numerous reference data and the other part of them have cumbersome preprocessing steps to divide the character into different parts to be learned and transferred separately. In this paper, we propose a simple but powerful end-to-end Chinese calligraphy font generation framework ZiGAN, which does not require any manual operation or redundant preprocessing to generate fine-grained target style characters with few-shot references. To be specific, a few paired samples from different character styles are leveraged to attain fine-grained correlation between structures underlying different glyphs. To capture valuable style knowledge in target and strengthen the coarse-grained understanding of character content, we utilize multiple unpaired samples to align the feature distributions belonging to different character styles. By doing so, only a few target Chinese calligraphy characters are needed to generated expected style transferred characters. Experiments demonstrate that our method has a state-of-the-art generalization ability in few-shot Chinese character style transfer.
\end{abstract}


\maketitle

\section{Introduction}
Chinese characters are an ancient and precious cultural heritage. In China, Chinese characters are called `zi'. Since ancient times, countless outstanding calligraphers have left their valuable handwritings, which have become the brilliant achievements of human civilization. However, many valuable calligraphy works have been lost in the long history~\cite{tseng1993history}. Unlike English, which has only 26 letters, there are tens of thousands of characters in Chinese characters, each of which has a different glyph and represents a different meaning. Furthermore, different calligraphers have their own writing styles with special overall structure and stroke details. Therefore, it is very meaningful and challenging to generate a complete personalized font library with only a few references.
\begin{figure}
\centering
\includegraphics[scale=0.34]{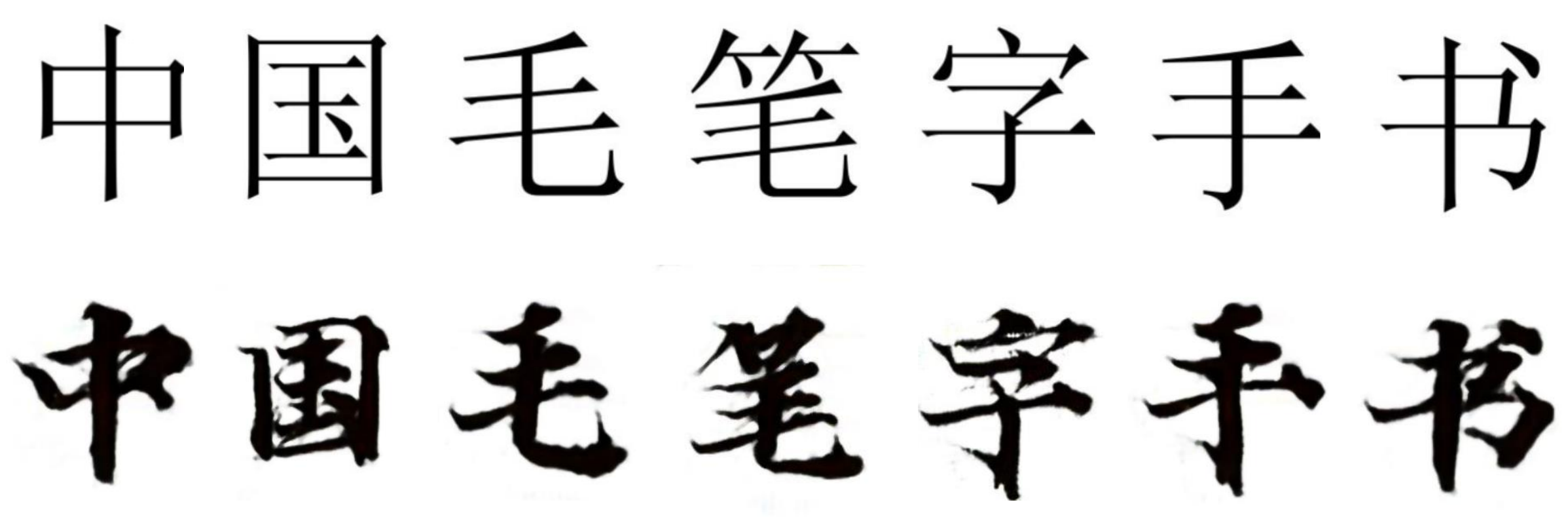}
\caption{The upper row is the input standard characters in font style Song, and the following line is the generated characters with target style. The generated calligraphy font shown in the figure has been successfully implemented in the application scenario.}
\label{fig:nsh}
\vspace{-2em}
\end{figure}

Some specific fonts have a relatively complete font library, for example, the widely used Chinese font Sim Sun version 5.16 covers 28,762 Unicode characters. But for most calligraphy works, it is almost impossible to get enough authentic works. The automatic generation of glyph images can greatly reduce the labor cost of font designers. Meanwhile, it is very helpful for calligraphy beginners to emulate the masterpieces reproduced.


Early studies on Chinese character synthesis tend to decompose characters into different radicals and regions, and then reassemble them~\cite{xu2005automatic,xu2009automatic}. But this kind of methods requires a lot of manual intervention and is inefficient. Additionally, they still produce undesirable results.

With the development of deep learning and computer vision, style transfer is discovered~\cite{gatys2016image,johnson2016perceptual,ulyanov2016texture,huang2017arbitrary}, which is dedicated to transforming one style of artwork into another. It achieves success in texture features transfer tasks, but unable to adapt to the translation in large geometric variations. Subsequently, methods such as pix2pix~\cite{pix2pix} and CycleGAN~\cite{cycleGAN} are proposed to solve image-to-image translation problem. But unlike photo-to-artwork task, Chinese characters are made up of pure black and white. More importantly, any lack of subtle structure or changes is unacceptable, while the GAN-based methods~\cite{yi2017dualgan,UGATIT,starGAN,starganv2} often lead to minor inaccuracy or blur.

Recently, some studies have been conducted to generate fonts~\cite{upchurch2016z,lyu2017auto,DCfont}. Zi2zi~\cite{zi2zi} is proposed based on the pix2pix framework, which results in good synthesizing performance in some specific font styles. On this basis, CalliGAN~\cite{wu2020calligan} further uses the prior information of Chinese character radicals to achieve better results. But this leads to a more complex and fragmented network structure. Furthermore, ChiroGAN~\cite{ChiroGAN} 
is committed to getting reasonable results without using paired data. But it cannot handle brush-written calligraphy images with complex skeletons. Moreover, all the abovementioned methods require a large number of style reference glyphs to achieve acceptable results, which may be laborious or even impossible to obtain in real-world scenarios. RD-GAN~\cite{RDGAN} 
is committed to using only a few style references, but it still requires a lot of prior knowledge of radicals, which will be very troublesome to process. And it generates handwritten photo-style images, which is different from our calligraphy written in ink on a white background.

In this paper, we propose ZiGAN, a novel end-to-end framework for fine-grained Chinese calligraphy font generation with few-shot target references. Given a few calligrapher's characters of the expected style, we can easily obtain the corresponding standard font images of the same characters and get the well-aligned pairs. We leverage these small amounts of paired samples to attain fine-grained correlation between structures underlying different styles.

Meanwhile, brush-written calligraphic character images are much more irregular than font-rendered character images. Few existing papers use this type of images to conduct experiments. In order to deal with this situation, we pioneer the utilization of numerous other unpaired characters in the standard font library which can be easily rendered. Although the glyphs of these characters are different from the target, they contain rich structure and morphological information. To capture valuable style knowledge in target and strengthen the coarse-grained understanding of character content, we utilize multiple unpaired samples to align the feature distributions belonging to different character styles. Figure \ref{fig:nsh} shows a successful application case of our method.

To sum up, our major contributions are summarized as follows: 

(1) We propose a simple but effective end-to-end framework that can generate fine-grained stylized calligraphy characters with only a few references. And it can easily adapt to a new handwriting style transfer task without tedious manual operations or prior knowledge.

(2) We innovatively learn the coarse-grained content knowledge of unpaired characters in the standard font library. To capture valuable structural knowledge, we map the features of the characters in different styles to Hilbert space and align the feature distributions. By doing so, we not only retain the semantic information of the character but also successfully translate the style from source to target while only a few target Chinese calligraphy characters are needed.

(3) Comprehensive experiments and analysis show that our approach can generate Chinese characters with state-of-the-art quality. More importantly, our method has been successfully implemented in actual application scenarios.
    \section{Related Work}
\subsection{Generative Adversarial Networks}
Generative Adversarial Networks (GAN)~\cite{GAN} has attracted a lot of interest since it was proposed. It has been successfully applied in many different fields and achieved impressive results, such as image generation~\cite{martin2017wasserstein,karras2017progressive,karras2019style}, image completion~\cite{iizuka2017globally,yu2018generative,yu2019free}, image editing~\cite{zhu2016generative}, transfer learning~\cite{DICD,JADA}, image translation~\cite{cycleGAN,pix2pix,UGATIT,starGAN}, etc. The key to the success of GAN is that the discriminator tries to distinguish the generated 
images from the realistic images, while the generator tries to confuse the judgment of the discriminator. In this paper, our model is based on GAN and only uses a few reference data to learn the Chinese calligraphy character style translation.

\subsection{Image-to-Image Translation}
Image-to-image translation aims to learn a mapping function that can transform an image from the source domain to the target domain. It has been widely used in many applications, for example, for artistic style transfer~\cite{johnson2016perceptual,chen2016fast}, semantic segmentation~\cite{long2015fully,noh2015learning,SSAN}, photo enhancement or object replacement.

A great quantity of GAN-based methods have been proposed, quite a few of them condition on images~\cite{pix2pix,starGAN,starganv2,wang2018high}. Pix2pix~\cite{pix2pix} is the pioneering method to figure out image-to-image translation. It follows the idea of conditional GAN, applying adversarial loss and L1-loss, and achieves impressive results. After that, high-resolution version is proposed to reinforce pix2pix in image synthesis and semantic manipulation~\cite{wang2018high}. But those paired training data are hard to obtain for some applications such as artistic style transfer. To alleviate this pain point, unpaired image-to-image translation frameworks have been proposed where no paired data are available anymore~\cite{cycleGAN,liu2017unsupervised,liu2016coupled}. It is a remarkable fact that CycleGAN~\cite{cycleGAN} proposes the cycle-consistent adversarial network, where two GANs interact in a cycle and learn source and target image distributions simultaneously. Based on CycleGAN, U-GAT-IT~\cite{UGATIT}  proposes a novel method for unsupervised image-to-image translation, which incorporates a new attention module and a new learnable normalization function called AdaLIN in an end-to-end manner. It is effective in the task of animating faces. In summary, the aforementioned paired methods all require a lot of data for training, otherwise the results will be unsatisfactory. Meanwhile, the unpaired methods often cause missing or redundant construction. But in the task of Chinese calligraphy character style translation, calligraphers’ handwriting is often difficult to obtain, and we cannot tolerate the inconsistency of character structure.

\subsection{Chinese Font Generation}
Chinese font generation has been studied for a long time~\cite{wong2000virtual,wu2006web}. The image-based methods ~\cite{xu2005automatic,du2016bayesian} split and reorganize the corresponding strokes and radicals in the dataset to generate the characters we want. But these methods contain too much human intervention, which is very inconvenient. With the development of deep learning, people have paid more attention to GAN-based character translation. Since character translation requires higher accuracy according to its complex strokes and style, it is more difficult than classic image-to-image translation problems. Transferring the styles of the alphabet is quite helpful and efficient for English translation~\cite{azadi2018multi}. While it is not simple like this for Chinese character style transfer because each Chinese character has its own glyph shape and there is a large number of existed characters. The style of the strokes in a certain character may quite different from the same strokes in other characters~\cite{strokeRefinement}, which makes the problem harder.

The first way to generate Chinese characters is following image-to-image methods, like zi2zi~\cite{zi2zi}, an open-source project that was never published as a paper. It's based on pix2pix, trying to translate character images from source style to various target styles. Based on zi2zi, DCFont~\cite{DCfont} and PEGAN~\cite{sun2018pyramid} have made improvements and achieved better results.
The second way to synthesize Chinese characters often separates a character into two parts, which are content and style~\cite{ChiroGAN,strokeRefinement}. EMD~\cite{EMD} and SA-VAE~\cite{SAVAE} use two different encoders to process content and style respectively. After absorbing the advantages of the above methods, CalliGAN~\cite{wu2020calligan} adds an extra component code of the character to train a conditional GAN, exploiting prior knowledge to maintain the structure information. While it needs a dictionary for each Chinese character to save its component code, this is a complicated preprocessing work. Unlike the aforementioned methods, ChiroGAN~\cite{ChiroGAN} uses erosion and dilation operations to obtain the basic skeleton of characters, then transfers style from source to target at the skeleton level. The output of this module is the skeleton image so it has to use another network to render the skeleton into the target character. Moreover, it relies on the effects of corrosion and expansion algorithms so that it often crashes on complex characters with numerous strokes or irregular glyph styles.

In order to save the cost of multiple Chinese characters selection, several recent methods aim to generate new glyphs with few numbers style references. DMfont~\cite{DMfont} disassembles Korean or
Thai glyphs to stylize components and then reassembles them. But it cannot handle complex Chinese characters. RD-GAN~\cite{RDGAN} aims to generate unseen characters in the fixed style, but it still requires a lot of prior knowledge of radicals, which will be very troublesome to process. Other earlier few-shot methods also have fatal shortcomings, such as being unable to generate complex glyphs~\cite{azadi2018multi} or failing to capture local styles~\cite{SAVAE,gao2019artistic}.

To sum up, part of the methods require lots of data, but the handwritings of many ancient Chinese calligraphers are not handed down so we cannot obtain them. The other parts of the methods are doped with too much manual processing. Moreover, they utilize too much intricate prior knowledge, which makes the preprocessing work complicated and can only adapt to a single task. To overcome these challenges, in this paper we propose a novel ZiGAN that can learn an intact and delicate
character style and structure when only a few target characters are provided. ZiGAN is an end-to-end framework, which can be easily and conveniently applied to any character style translation task, and is capable of generating a complete and consistent font library.

\begin{figure*}[ht]
\centering
\includegraphics[scale=0.55,height=10cm]{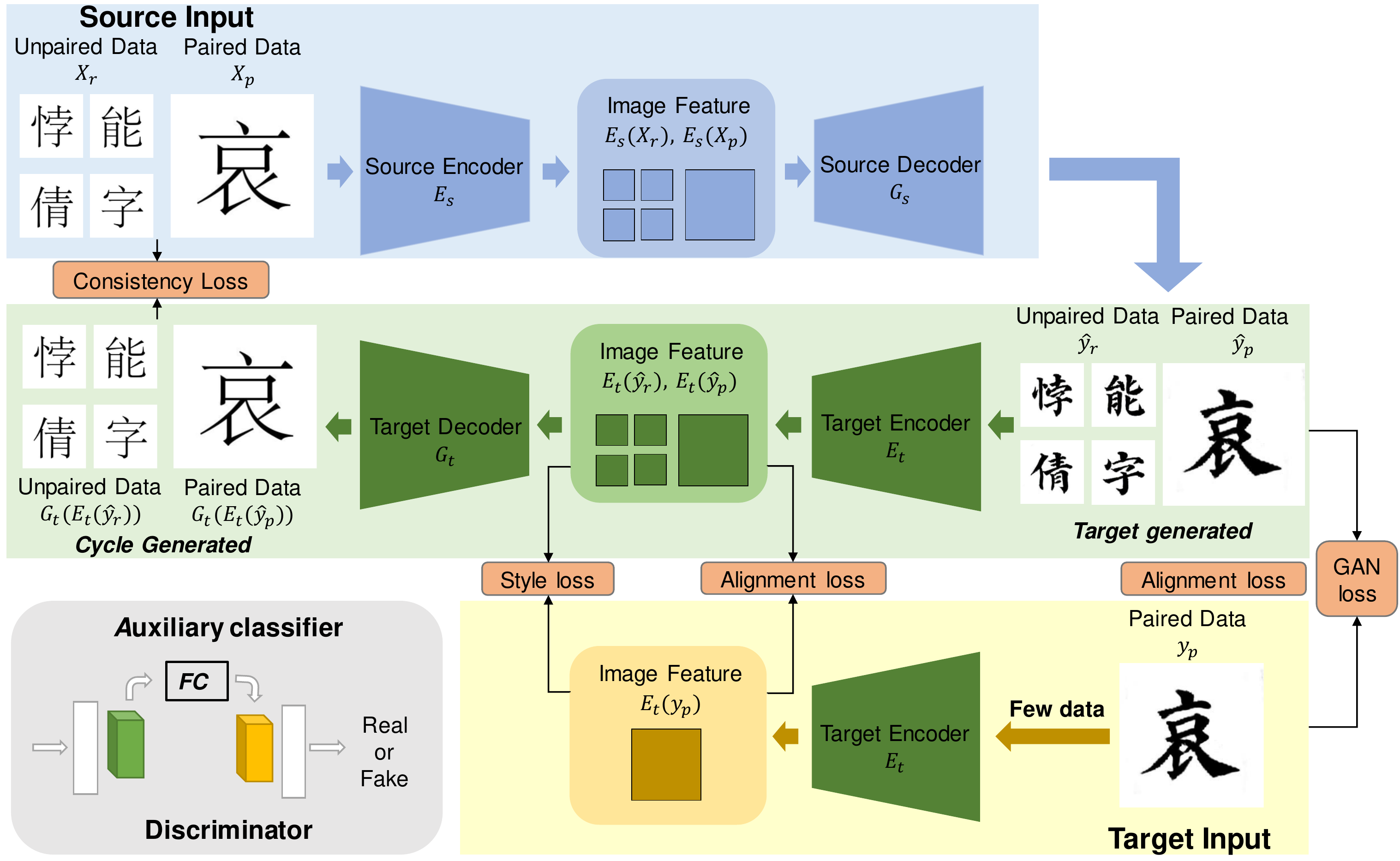}
\caption{Network architectures. ZiGAN is an end-to-end framework based on the encoder and decoder. The network can not only learn style information from a few target images but also learn structure and content information from numerous source images. An auxiliary classifier is added to the discriminator to force the model to focus on more important regions. ZiGAN has 4 losses: GAN loss (Eq.~\eqref{eq:GAN}), consistency loss (Eq.~\eqref{eq:consistency}), alignment loss (Eq.~\eqref{eq:alignment}), style loss (Eq.~\eqref{eq:style}).}
\label{fig:network achitecture}
\end{figure*}

\section{Method}
We distinguish each Chinese character based on structure, radicals and strokes. Therefore, each calligrapher writes the same content of Chinese characters but in different styles. The goal of our proposed method is to learn a way to generate Chinese character images with the expected style from only a small amount of given characters. Let $s$ be the style we want, and $y$ be a target image under the style $s$. We use TrueType fonts to render a source image $x_{p}$ representing the same character as $y$ in black with font style Song as the standard character image. Furthermore, we find that although we can only obtain a few target character images, we can render a mass of source character images from the TTF of font style Song. We randomly render images from font style Song, defined as $x_{r}$. In general, we leverage the paired image sets \{$x_{p}$\} and \{$y$\} to attain fine-grained correlation between structures underlying different glyphs. Moreover, the method we proposed learns the extra structural knowledge in the unpaired data \{$x_{r}$\} simultaneously to strengthen the coarse-grained understanding of the character content. Our framework consists of two generators and two discriminators in two opposite directions. Here we only explain the direction of $x\rightarrow{y}$ as the vice versa should be straightforward. 

\subsection{Network architectures}
Figure \ref{fig:network achitecture}  shows the architecture of our network.
We encode $x_{p}$ and $x_{r}$ into the feature space through an image encoder $E_{s}$, and then decode image features by an image decoder $G_{s}$ to generate the stylized character images $\hat{y}_{p}$ and $\hat{y}_{r}$. After that, we model on CycleGAN~\cite{cycleGAN} and set up a reversing generator, including encoder $E_{t}$ and decoder $G_{t}$.

\noindent\textbf{Image encoder and decoder.} We use the encoder-decoder architecture as our generator, which is based on pix2pix~\cite{pix2pix} and zi2zi~\cite{zi2zi} with some improvements. Unlike zi2zi, we remove the category embedding vector because it is inapplicable for our task and will increase instability. The complete architecture of our generator is in Table \ref{table:generator}. All convolution and deconvolution layers use 5-by-5 filters with stride size of 2, and apply batch normalization. The encoder layers actually use LeakyReLU for activation function with a slope of 2. While the decoder layers use the activation function ReLU. We use dropout with a rate of 0.5 only in L1 to L3 layers of the decoder.

\begin{table}
    \centering
    \caption{The architecture of encoder and decoder.}
    \label{table:generator}
    \setlength{\tabcolsep}{5mm}{
    \renewcommand\arraystretch{1}
    \small
    \begin{tabular}{ccc}
    \hline Layer & Encoder & Decoder \\
    \hline Input & $256 \times 256 \times 3$ & $1 \times 1 \times 512$ \\
    L1 & $128 \times 128 \times 64$ & $2 \times 2 \times 1024$ \\
    L2 & $64 \times 64 \times 128$ & $4 \times 4 \times 1024$ \\
    L3 & $32 \times 32 \times 256$ & $8 \times 8 \times 1024$ \\
    L4 & $16 \times 16 \times 512$ & $16 \times 16 \times 1024$ \\
    L5 & $8 \times 8 \times 512$ & $32 \times 32 \times 512$ \\
    L6 & $4 \times 4 \times 512$ & $64 \times 64 \times 256$ \\
    L7 & $2 \times 2 \times 512$ & $128 \times 128 \times 128$ \\
    L8 & $1 \times 1 \times 512$ & $256 \times 256 \times 3$ \\
    \hline
    \end{tabular}}
    \vspace{-1.5em}
\end{table}

\noindent\textbf{CAM Discriminator.} We add an auxiliary classifier $\eta_{D_t}$ based on Class Activation Map (CAM)~\cite{CAM} to the discriminator so that the model can pay more attention to more important regions. For different calligraphy, there may be subtle but critical differences between the strokes and radicals. The local and global discriminator with CAM attention module can help the model distinguish better and generate finer characters of different styles. Unlike pix2pix~\cite{pix2pix} and zi2zi~\cite{zi2zi}, we don't use conditional image knowledge to reduce complexity so the discriminator does not observe $x$. In Section \ref{section4}, we demonstrate that the CAM attention module can learn the details successfully.
\subsection{Loss Function}
We define four losses in total. The loss items of $x\rightarrow{y}$ can be written as:

\textbf{GAN loss.} GAN loss is divided into main and auxiliary parts. In the main part, we impose adversarial loss to match the distribution of the translated images and target images. We use the Least Squares GAN~\cite{lsGAN} objective to train our model.
\begin{equation}
\begin{aligned}
\mathcal{L}_{adv}^{x \rightarrow y}=& \mathbb{E}_{y}[(D_t(y))^2]+\\
& \mathbb{E}_{x}[(1-D_t(G_s(E_s(x))))^2].
\end{aligned}
\end{equation} 

In addition, we add an auxiliary classifier $\eta_{D_t}$ based on Class Activation Map(CAM)~\cite{CAM} to the discriminator $D_t$. Let $y {\in\left\{Y\}, G_s(E_s\left(X\right)\right)}$ represent a sample from the target domain and the translated source domain. The discriminator $D_t$ consists of an encoder $E_{D_t}$, a classifier $C_{D_t}$, and an auxiliary classifier $\eta_{D_t}$. The auxiliary classifier is trained to learn the weight of the $k$-th feature map for the target domain, $w_{t}^{k}$, by using the global average pooling and global max pooling, i.e., ${\eta_{D_t}(y)=\sigma\left(\Sigma_{k} w_{t}^{k} \Sigma_{i j} E_{D_t}^{k_{i j}}(y)\right)}$. By exploiting $w_{t}^{k}$, we can calculate a set of domain specific attention feature map $a_{D_t}(y)=w_{D_t} * E_{D_t}(y)=\left\{w_{D_t}^{k} *E_{D_t}^{k}(y) \mid 1 \leq k \leq n\right\}$, where $n$ is the number of encoded feature maps. Then, our discriminator $D_{t}(y)$ becomes equal to $C_{D_t}(a_{D_t}(y))$. By doing so, the discriminator can better distinguish the differences in the details of different character styles, while $E_s$ and $G_s$ can make improvements in the most important regions.
\begin{equation}
\begin{aligned}
\mathcal{L}_{cam}^{x \rightarrow y}=& \mathbb{E}_{y}[( \eta_{D_t}(y))^2]+\\
& \mathbb{E}_{x}[(1-\eta_{D_t}(G_s(E_s(x)))^2].
\end{aligned}
\end{equation} 
On the whole:
\begin{equation}
\begin{aligned}
\mathcal{L}_{GAN}^{x \rightarrow y}=\mathcal{L}_{adv}^{x \rightarrow y}+\mathcal{L}_{cam}^{x \rightarrow y}.
\label{eq:GAN}
\end{aligned}
\end{equation} 

\textbf{Consistency loss.} We constrain the consistency of the model from two parts. First, the model must have the ability to cycle back. It means that after $x$ is translated to $\hat{y}$, it must be successfully translated back to the original domain:
\begin{equation}
\begin{aligned}
\mathcal{L}_{cycle}^{x \rightarrow y}=& \mathbb{E}_{x}[|{x-G_t(E_t(G_s(E_s(x))))}|_1].
\end{aligned}
\end{equation} 
Second, identity loss is used to constrain the color and shape of the characters to not be distorted. Given an image $y$, after the translation of $Es$ and $Gs$, it should be the same character in the same style.
\begin{equation}
\begin{aligned}
\mathcal{L}_{identity}^{x \rightarrow y}=& \mathbb{E}_{y}[|{y-(G_s(E_s(y)))}|_1].
\end{aligned}
\end{equation} 
So the total consistency loss is:
\begin{equation}
\begin{aligned}
\mathcal{L}_{consistency}^{x \rightarrow y}=\mathcal{L}_{cycle}^{x \rightarrow y}+\mathcal{L}_{identity}^{x \rightarrow y}.
\label{eq:consistency}
\end{aligned}
\end{equation} 

\textbf{Alignment loss } We align the content and feature levels of $x_p$ and $y$ to leverage the paired samples to attain fine-grained structural correspondence. In the font style translation task, the job of the discriminator is still to distinguish which is generated or which is real, but the generator is tasked to not only fool the discriminator, but also to be as similar to the ground truth at the content level as possible. We use the L1 loss to constrain the output of paired data $x_p$,
\begin{equation}
\begin{aligned}
\mathcal{L}_{L1}^{x \rightarrow y}=& \mathbb{E}_{x,y}[|{y-G_s(E_s(x_p))}|_1].
\end{aligned}
\end{equation} 
And in order to constrain the features of the generated image and the real image to the same space, we apply constancy loss:
\begin{equation}
\begin{aligned}
\mathcal{L}_{constancy}^{x \rightarrow y}=& \mathbb{E}_{x,y}[|{E_t(y)-E_t(G_s(E_s(x_p)))}|_2].
\end{aligned}
\end{equation} 
Therefore, the total alignment loss can be formulated as:
\begin{equation}
\begin{aligned}
\mathcal{L}_{alignment}^{x \rightarrow y}=\alpha\mathcal{L}_{L1}^{x \rightarrow y}+\mathcal{L}_{constancy}^{x \rightarrow y}.
\label{eq:alignment}
\end{aligned}
\end{equation} 
where $\alpha=5$.

\textbf{Style loss } For better understanding of coarse-grained character content and a maturer style translation, we have introduced style loss to take advantage of multiple unpaired samples $x_r$. Unlike paired data, unpaired data cannot simply be restricted by L1 or L2 losses. Therefore, with comprehensive consideration of time complexity and computational cost, we utilize MK-MMD~\cite{gretton2007kernel,DRCNPDA} to match the feature distributions to retain style information. Denote by $\mathcal{H}_{k}$ be the reproducing kernel Hilbert space (RKHS) endowed with a characteristic kernel $k$. The mean embedding of distribution $p$ in $\mathcal{H}_{k}$ is a unique element $\mu_{k}(p)$ such that $\mathbf{E}_{\mathbf{x} \sim p} f(\mathbf{x})= \left\langle f(\mathbf{x}), \mu_{k}(p)\right\rangle_{\mathcal{H}_{k}}$ for all $f$ $\in \mathcal{H}_{k}$. The MK-MMD $d_k(x,y)$ between probability distributions $x$ and $y$ is defined as the RKHS distance between the mean embeddings of $x$ and $y$. The squared formulation of style loss is defined as:
\begin{equation}
\begin{aligned}
\mathcal{L}_{style}^{2}=&||\mathbb{E}_{y}[{\phi(E_t(y))]-\mathbb{E}_{x}[\phi(E_t(G_s(E_s(x_r))))}]||_{\mathcal{H}_{k}}^2.
\label{eq:style}
\end{aligned}
\end{equation} 
where $\phi$ is the corresponding feature map. And it’s worth noting that when $x=y$, $\mathcal{L}_{style}$=0.
Here we choose Gaussian kernel function as the kernel function:
\begin{equation}
\begin{aligned}
k_{\sigma}^{r b f}(P_s, P_t)=\exp \left(-\frac{1}{2 \sigma^{2}}\|x-y\|^{2}\right).
\end{aligned}
\end{equation}
\textbf{Full objective } Finally, the full objective function is defined as:
\begin{equation}
\begin{aligned}
\begin{aligned}
\mathcal{L} &=\lambda_{1} \mathcal{L}_{GAN}+\lambda_{2} \mathcal{L}_{consistency}+\lambda_{3} \mathcal{L}_{alignment} \\
&+\lambda_{4} \mathcal{L}_{style}.
\end{aligned}
\end{aligned}
\end{equation} 
where $\lambda_{1}=5$,~$\lambda_{2}=10$,~$\lambda_{3}=10$,~$\lambda_{4}=10$. Here $\mathcal{L}_{GAN}$ = $\mathcal{L}_{GAN}^{x \rightarrow y}$ + $\mathcal{L}_{GAN}^{y \rightarrow x}$ and the other losses are defined in the similar way.
\section{Experiment}\label{section4}
\setlength{\textfloatsep}{5pt}
\subsection{Datasets}

\begin{figure}
    \centering
    \setlength{\tabcolsep}{0.2mm}{
    \begin{tabular}{ccccccccc}
    \includegraphics[scale=0.1]{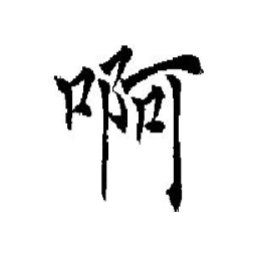}
    & \includegraphics[scale=0.1]{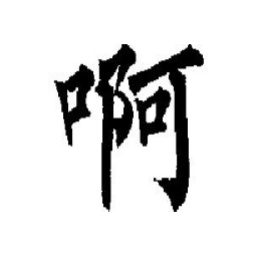}
    & \includegraphics[scale=0.1]{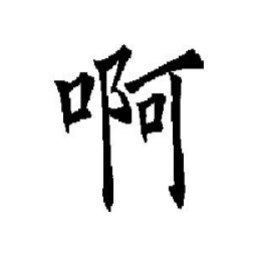}
    & \includegraphics[scale=0.1]{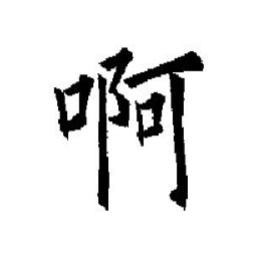}
    & \includegraphics[scale=0.1]{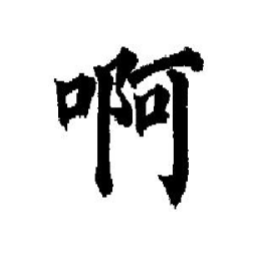}
    & \includegraphics[scale=0.1]{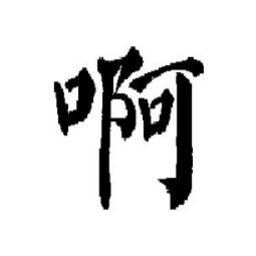}
    & \includegraphics[scale=0.1]{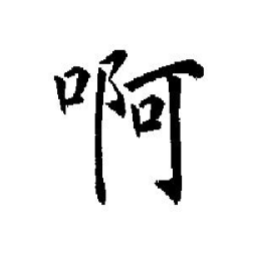}
    & \includegraphics[scale=0.1]{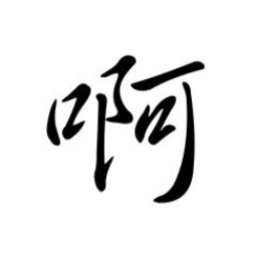}
    & \includegraphics[scale=0.1]{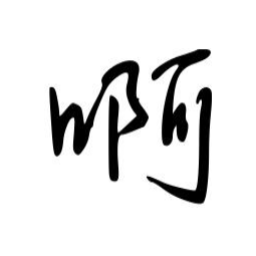}\\
    \end{tabular}}
    \caption{Example characters in 9 different styles.(1-9 in order)}
    \label{fig:data example.}
\end{figure}

\begin{table}
    \centering
    \setlength{\tabcolsep}{3mm}
    \caption{Data sets used in our experiments. Dataset Name is the expert of the masterpiece and the sub-typeface for those created with the same master. Samples is the number of characters in each data set.}
    \small
    \begin{tabular}{ c | p{4.5cm}<{\centering} | c }
        \hline
            Style & Dataset Name & Samples \\
            \hline
            1 & \multicolumn{1}{l|}{Chu Suiliang} & 7159 \\
            2 & \multicolumn{1}{l|}{Liu Gongquan} & 6171 \\
            3 & \multicolumn{1}{l|}{Ouyang Xun} & 6999 \\
              & \multicolumn{1}{l|}{\quad \small{— Huangfu Dan Stele}} &  \\
            4 & \multicolumn{1}{l|}{Ouyang Xun} & 6901 \\
              & \multicolumn{1}{l|}{\quad \small{— Inscription on Sweet Wine}} &  \\
              & \multicolumn{1}{l|}{\qquad \small{ Spring at Jiucheng Palace}} &  \\
            5 & \multicolumn{1}{l|}{Yan Zhenqing} & 6308 \\
              & \multicolumn{1}{l|}{\quad \small{— Stele of the Abundant}} &  \\
              & \multicolumn{1}{l|}{\qquad \small{ Treasure Pagoda}} &  \\
            6 & \multicolumn{1}{l|}{Yan Zhenqing} & 7006 \\
              & \multicolumn{1}{l|}{\quad \small{— Yan Qinli Stele}} &  \\
            7 & \multicolumn{1}{l|}{Yu Shinan} & 7008 \\ 
            8 & \multicolumn{1}{l|}{XING} & 6800 \\ 
            9 & \multicolumn{1}{l|}{CAO} & 6799 \\ 
            \hline
    \end{tabular}
\label{table:dataset}
\end{table}

To better show our model's performance, we use the same datasets with CalliGAN~\cite{wu2020calligan}. The datasets could be downloaded from a Chinese calligraphy character website\footnote{\url{http://163.20.160.14/~word/modules/myalbum/}}, where there are more than 20 kinds of brush-written calligraphy sets belonging to different Chinese ancient experts. And 7 styles belonging to regular script are used to complete our experiment. The 3rd and the 4th style sets are the same ancient calligraphic expert's masterpieces created in different periods of his life.
They are treated as two different style sets due to the differences between them, which is also the rule of thumb in the Chinese calligraphy community.
The 5th and the 6th style sets are the same situation as above. In addition to the above-mentioned dataset which is the same as CalliGAN, we also test our model in other more irregular and challenging Chinese character fonts like XING and CAO to prove that our method is highly adaptable and robust to any font style. So our data set consists of 9 fonts in total which are shown in Table \ref{table:dataset}. Figure \ref{fig:data example.} shows example characters in 9 different styles.

\begin{figure*}
    \centering
    \begin{tabular}{ c }
        {\includegraphics[scale=0.89]{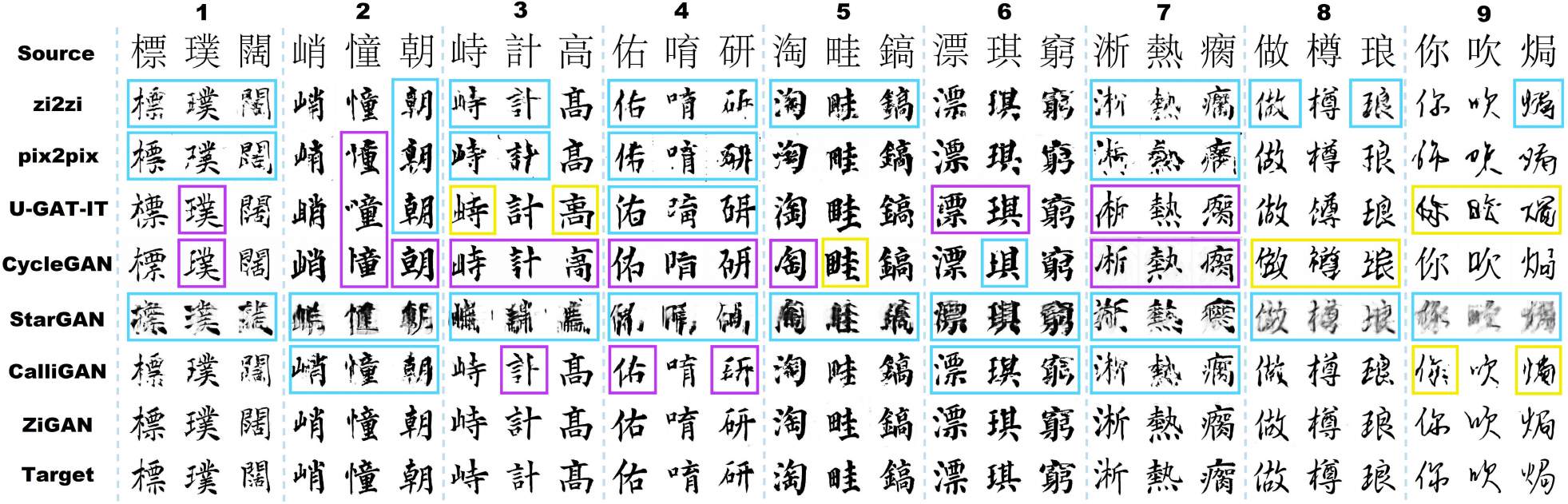}} \\
    \end{tabular}
    ~\\
    \caption{Comparison of the results using each method in 9 different font styles. Characters in the purple box are generated with missing strokes, the yellow box means incorrect extra strokes translation, and the blue box indicates conspicuous blurred results.}
    \label{fig:conparison}
\end{figure*}

We collect 61151 target images that cover 6560 characters in the 9 styles in all. And we use TTF of font style Song to render source image $x$. To explore the ability of our few-shot method, we create two configurations for the dataset: 100-shot and 200-shot. Each style has 100 or 200 randomly selected training examples respectively as input {$y$}, while the remaining images as the test set. Such a training set size is much smaller than other methods which often require thousands of training images. Specific information is listed in Table \ref{table:statistics}. Given ${y}$, we can easily get the same character image ${x_p}$ in the source domain. In the meantime, we randomly sample and render 6000 unpaired images with font style Song which cover a large number of characters as input $x_r$.

The images in this repository have various shapes depending on the character's shapes. We follow the preprocessing steps of CalliGAN~\cite{wu2020calligan}, but the only difference is that we process the images into three-channel RGB images. So we get $256\times256\times3$ images as our ground truth $y$. All images are converted to tensors linearly with a value range between -1 and 1 by our network.

\begin{table}[!t]
    \centering
    \setlength{\tabcolsep}{1mm}{
    \caption{Statistics of our 100-shot and 200-shot configurations.}
    \label{table:statistics}
    \small
    \begin{tabular}{c|ccccccccc}
    \hline Style & 1 & 2 & 3 & 4 & 5 & 6 & 7 & 8 & 9 \\
    \hline
    \multicolumn{1}{l|}{100shot-Train} & 100 & 100 & 100 & 100 & 100 & 100 & 100 & 100 & 100 \\
    \multicolumn{1}{l|}{100shot-Test} & 7059 & 6071 & 6899 & 6801 & 6208 & 6906 & 6908 & 6700 & 6699 \\
    \hline
    \multicolumn{1}{l|}{200shot-Train} & 200 & 200 & 200 & 200 & 200 & 200 & 200 & 200 & 200 \\
    \multicolumn{1}{l|}{200shot-Test} & 6959 & 5971 & 6799 & 6701 & 6108 & 6806 & 6808 & 6600 & 6599 \\
    \hline Total & 7159 & 6171 & 6999 & 6901 & 6308 & 7006 & 7008 & 6800 & 6799 \\
    \hline
    \end{tabular}}
    ~\\
    ~\\
\end{table}

\begin{figure}[!h]
\centering
\includegraphics[scale=0.33]{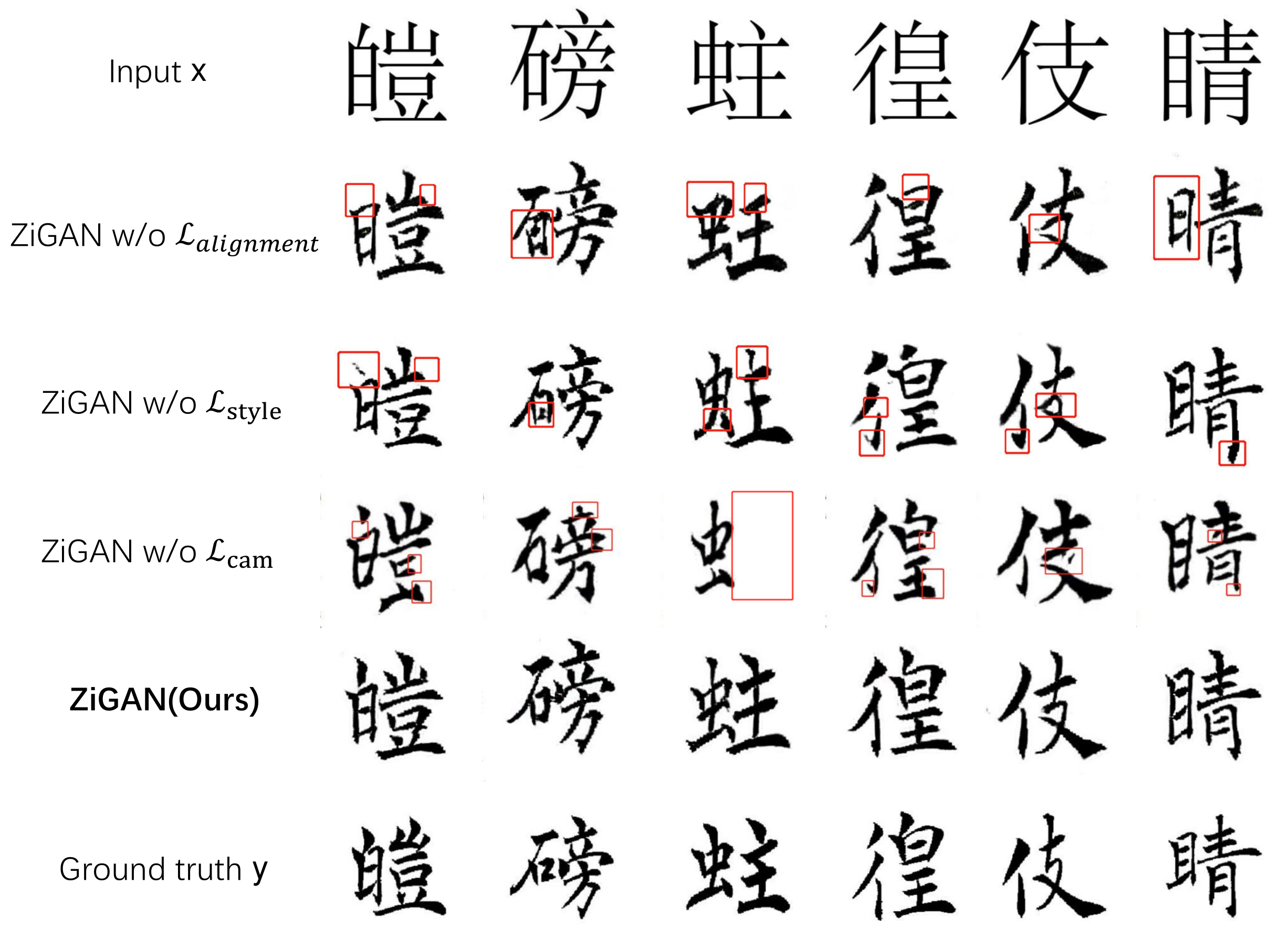}
\caption{The ablation experiment of ZiGAN. All characters are generated under style 1. The red rectangles mark the imperfect part of the character where some strokes are incomplete or fuzzy. It can be seen that every part of our method is beneficial to the result.}
\label{fig:xiaorong}
\end{figure}

\subsection{Experiment  Setup}
We use standard methods to optimize our network: first optimize on D, then on E and G together. Similarly, we also alternate training on $x_r$ and $x_p$. All models are trained using Adam with $\beta_1=0.5$, $\beta_2=0.999$. The learning rate is initialized to $0.0003$ and drops by half every $500$ epochs. Because the number of training samples is small, we train our model in $1500$ epochs.

\begin{table*}[]
    \centering
    \setlength{\tabcolsep}{2.3mm}{
    \renewcommand\arraystretch{1}
    \renewcommand{\multirowsetup}{\centering}
    \caption{IOU for difference font style translation mode. Higher is better. The seventh and eighth methods show the results of ablation experiments without $\mathcal{L}_{style}$ or $\mathcal{L}_{alignment}$.}
    \vspace{-1em}
    \label{tab:IOU}
    \small
    \begin{tabular}{m{6mm}|c|m{9.2mm}<{\centering}|m{9.2mm}<{\centering}|m{9.2mm}<{\centering}|m{9.2mm}<{\centering}|m{9.2mm}<{\centering}|m{9.2mm}<{\centering}|m{9.2mm}<{\centering}|m{9.2mm}<{\centering}|m{9.2mm}<{\centering}|m{9.2mm}<{\centering}}
        \hline
        \multicolumn{12}{c}{\textbf{Intersection Over Union ~ (IOU)}} \\
        \hline
        \multirow{2}{*}{Type} & \multirow{2}{*}{Method}
        & \multicolumn{10}{c}{Style} \\
        \cline{3-12}
        & & 1 & 2 & 3 & 4 & 5 & 6 & 7 & 8 & 9 & Mean \\
        \hline
        \multirow{9}{*}{\tabincell{c}{100 \\ - \\ shot}}
        & zi2zi & 0.228 & 0.353 & 0.282 & 0.298 & 0.381 & 0.37 & 0.274 & 0.289 & 0.213 & 0.299\\
        & pix2pix & 0.259 & 0.374 & 0.303 & 0.314 & 0.416 & 0.404 & 0.281 & 0.305 & 0.21 & 0.318\\
        & U-GAT-IT & 0.234 & 0.315 & 0.261 & 0.264 & 0.346 & 0.342 & 0.243 & 0.292 & 0.211 & 0.279\\
        & CycleGAN & 0.24 & 0.241 & 0.259 & 0.28 & 0.379 & 0.369 & 0.254 & 0.26 & 0.2 & 0.276\\
        & StarGAN & 0.152 & 0.3 & 0.194 & 0.206 & 0.373 & 0.314 & 0.151 & 0.21 & 0.195 & 0.233\\
        & CalliGAN & 0.241 & 0.345 & 0.277 & 0.293 & 0.382 & 0.391 & 0.278 & 0.306 & 0.218 & 0.303\\
        & ZiGAN w/o $\mathcal{L}_{style}$ & 0.236 & \textbf{0.402} & 0.308 & 0.332 & \textbf{0.417} & 0.403 & 0.291 & 0.313 & 0.203 & 0.323\\
        & ZiGAN w/o $\mathcal{L}_{align}$ & 0.229 & 0.326 & 0.291 & 0.309 & 0.383 & 0.377 & 0.272 & 0.296 & 0.19 & 0.297\\
        & \textbf{ZiGAN(Ours)} & \textbf{0.273} & 0.389 & \textbf{0.317} & \textbf{0.333} & 0.408 & \textbf{0.413} & \textbf{0.293} & \textbf{0.334} & \textbf{0.226} & \textbf{0.332}\\
        \hline
        \multirow{9}{*}{\tabincell{c}{200 \\ - \\ shot}}
        & zi2zi & 0.257 & 0.395 & 0.308 & 0.324 & 0.426 & 0.407 & 0.274 & 0.319 & 0.233 & 0.327\\
        & pix2pix & 0.27 & 0.398 & \textbf{0.321} & 0.333 & 0.432 & 0.422 & 0.292 & 0.315 & 0.215 & 0.333\\
        & U-GAT-IT & 0.239 & 0.339 & 0.255 & 0.267 & 0.367 & 0.35 & 0.249 & 0.296 & 0.216 & 0.286\\
        & CycleGAN & 0.241 & 0.36 & 0.098 & 0.27 & 0.372 & 0.365 & 0.255 & 0.262 & 0.202 & 0.269\\
        & StarGAN & 0.2 & 0.331 & 0.26 & 0.221 & 0.374 & 0.359 & 0.21 & 0.221 & 0.235 & 0.268\\
        & CalliGAN & 0.267 & 0.347 & 0.319 & 0.324 & 0.414 & 0.404 & 0.289 & 0.327 & \textbf{0.236} & 0.325\\
        & ZiGAN w/o $\mathcal{L}_{style}$ & 0.286 & 0.397 & 0.312 & 0.332 & \textbf{0.438} & 0.425 & 0.295 & 0.323 & 0.233 & 0.338\\
        & ZiGAN w/o $\mathcal{L}_{align}$ & 0.284 & 0.379 & 0.302 & 0.303 & 0.394 & 0.403 & 0.287 & 0.333 & 0.23 & 0.324\\
        & \textbf{ZiGAN(Ours)} & \textbf{0.290} & \textbf{0.407} & 0.319 & \textbf{0.357} & 0.436 & \textbf{0.427} & \textbf{0.316} & \textbf{0.344} & \textbf{0.236} & \textbf{0.348}\\
        \hline
    \end{tabular}}
\end{table*}

\begin{table*}[]
    \centering
    \setlength{\tabcolsep}{2.3mm}{
    \renewcommand\arraystretch{1}
    \caption{The top-1 accuracy of generated characters. We train a Resnet18 model as a Chinese character recognizer on the ground truth of all styles. The recognition accuracy can show whether the characters retain the complete character structure and content. The seventh and eighth methods show the results of ablation experiments without $\mathcal{L}_{style}$ or $\mathcal{L}_{alignment}$.}
    \vspace{-1em}
    \label{tab:accuracy}
    \small
    \begin{tabular}{m{6mm}|c|m{9.2mm}<{\centering}|m{9.2mm}<{\centering}|m{9.2mm}<{\centering}|m{9.2mm}<{\centering}|m{9.2mm}<{\centering}|m{9.2mm}<{\centering}|m{9.2mm}<{\centering}|m{9.2mm}<{\centering}|m{9.2mm}<{\centering}|m{9.2mm}<{\centering}}
        \hline
        \multicolumn{12}{c}{\textbf{Top-1 Accuracy}} \\
        \hline
        \multirow{2}{*}{Type} & \multirow{2}{*}{Method}
        & \multicolumn{10}{c}{Style} \\
        \cline{3-12}
        & & 1 & 2 & 3 & 4 & 5 & 6 & 7 & 8 & 9 & Mean \\
        \hline
        \multirow{9}{*}{\tabincell{c}{100 \\ - \\ shot}}
        & zi2zi & 0.075 & 0.075 & 0.094 & 0.101 & 0.105 & 0.129 & 0.094 & 0.186 & 0.114 & 0.108\\
        & pix2pix & 0 & 0 & 0.048 & 0.054 & 0 & 0.123 & 0.068 & 0.196 & 0.1 & 0.065\\
        & U-GAT-IT & 0.156 & 0.025 & 0.156 & 0.139 & 0.128 & 0.2 & 0.166 & 0.102 & 0.11 & 0.131\\
        & CycleGAN & 0.204 & 0.064 & 0.215 & 0.245 & 0.351 & 0.314 & 0.351 & 0.13 & \textbf{0.16} & 0.226\\
        & StarGAN & 0.003 & 0.002 & 0.001 & 0.001 & 0.007 & 0.007 & 0.002 & 0.001 & 0 & 0.003\\
        & CalliGAN & 0.091 & 0.083 & 0.089 & 0.099 & 0.108 & 0.184 & 0.104 & 0.104 & 0.11 & 0.108\\
        & ZiGAN w/o $\mathcal{L}_{style}$ & 0.255 & 0.283 & 0.257 & 0.317 & 0.26 & 0.451 & 0.325 & 0.353 & 0.13 & 0.292\\
        & ZiGAN w/o $\mathcal{L}_{align}$ & 0.199 & 0.16 & 0.195 & 0.237 & 0.294 & 0.334 & 0.288 & 0.336 & 0.12 & 0.24\\
        & \textbf{ZiGAN(Ours)} & \textbf{0.567} & \textbf{0.740} & \textbf{0.647} & \textbf{0.625} & \textbf{0.592} & \textbf{0.733} & \textbf{0.694} & \textbf{0.404} & 0.158 & \textbf{0.573}\\
        \hline
        \multirow{9}{*}{\tabincell{c}{200 \\ - \\ shot}}
        & zi2zi & 0.175 & 0.346 & 0.207 & 0.227 & 0.323 & 0.326 & 0.196 & 0.246 & 0.118 & 0.24\\
        & pix2pix & 0.154 & 0.17 & 0.151 & 0.149 & 0.213 & 0.294 & 0.149 & 0.339 & 0.101 & 0.191\\
        & U-GAT-IT & 0.236 & 0.268 & 0.131 & 0.132 & 0.283 & 0.246 & 0.184 & 0.32 & 0.12 & 0.213\\
        & CycleGAN & 0.211 & 0.304 & 0.19 & 0.234 & 0.347 & 0.321 & 0.385 & 0.14 & 0.16 & 0.255\\
        & StarGAN & 0.005 & 0.012 & 0.001 & 0 & 0.013 & 0.017 & 0.005 & 0.004 & 0.002 & 0.007\\
        & CalliGAN & 0.192 & 0.169 & 0.26 & 0.212 & 0.276 & 0.353 & 0.23 & 0.334 & 0.117 & 0.238\\
        & ZiGAN w/o $\mathcal{L}_{style}$ & 0.586 & 0.614 & 0.528 & 0.577 & 0.62 & 0.637 & 0.593 & 0.403 & 0.16 & 0.462\\
        & ZiGAN w/o $\mathcal{L}_{align}$ & 0.528 & 0.583 & 0.568 & 0.347 & 0.566 & 0.24 & 0.582 & 0.396 & 0.146 & 0.44\\
        & \textbf{ZiGAN(Ours)} & \textbf{0.631} & \textbf{0.631} & \textbf{0.636} & \textbf{0.591} & \textbf{0.689} & \textbf{0.703} & \textbf{0.722} & \textbf{0.488} & \textbf{0.176} & \textbf{0.585}\\
        \hline
    \end{tabular}}
\end{table*}

\subsection{Qualitative Evaluation}
To prove the advancement of our method in the field of few-shot font style transfer, we have extensively compared various methods. Six classic methods are used as the baselines, including zi2zi~\cite{zi2zi}, pix2pix~\cite{pix2pix}, U-GAT-IT~\cite{UGATIT}, CycleGAN~\cite{cycleGAN}, StarGAN~\cite{starGAN}, CalliGAN~\cite{wu2020calligan}. Among them, zi2zi, pix2pix and CalliGAN need paired data for training. We use the same number of paired images as ours to train their model in corresponding configurations. CycleGAN, U-GAT-IT and StarGAN are unsupervised methods. We use 6200 images in font style Song as their source domain and 100 or 200 calligraphic images as their target domain for different configurations so that the size of their training set is not smaller than ours. Figure \ref{fig:conparison} shows the comparison of generation results.

CycleGAN not only did not fully learn the style of the characters but also lost some strokes. StarGAN has lost the structural information of the character and is completely unable to do this job. Pix2pix barely maintains the structure of the characters, but there are too many fuzzy and damaged places. U-GAT-IT seems to have learned the style of the font, but there are still many erroneous and missing strokes in the result. Although zi2zi and CalliGAN are professional in font style translation, they produce unsatisfactory results which contain too many meaningless blanks and blurs when few targets are referenced. Only ZiGAN has found the law of calligraphy from limited target characters.

\begin{table}[t]
    \centering
    \caption{User study. We ask respondents to choose which generated character resembles the ground truth from these 7 methods.}
    \vspace{-1em}
    \label{tab:User study}
    \small
    \begin{tabular}{m{15mm}<{\centering} m{18mm}<{\centering}|m{15mm}<{\centering} m{18mm}<{\centering} }
        \hline
        Method & Vote Rate(\%) & Method & Vote Rate(\%) \\
        \hline
        zi2zi & 1.52 & CycleGAN & 1.68  \\
        pix2pix & 1.44 & StarGAN & 0.31 \\
        U-GAT-IT & 2.27 & CalliGAN & 1.80 \\
        - & - & \textbf{ZiGAN(Ours)} & \textbf{90.98} \\
        \hline
    \end{tabular}
\end{table}

\begin{table}[ht]
    \centering
    \setlength{\tabcolsep}{1.85mm}{
    \renewcommand\arraystretch{2}
    \caption{Turing test samples. Each sample contains 6 fake glyph images generated by ZiGAN and 6 real glyph images. ZiGAN achieves an accuracy of 51.6\%, which is very close to random selection.}
    \vspace{-1em}
    \label{tab:Turing}
    \small
    \begin{tabular}{|c|c|c|}
        \hline
        Sample 1 & Sample 2 & Sample 3 \\
        \hline
        {\includegraphics[scale=0.093]{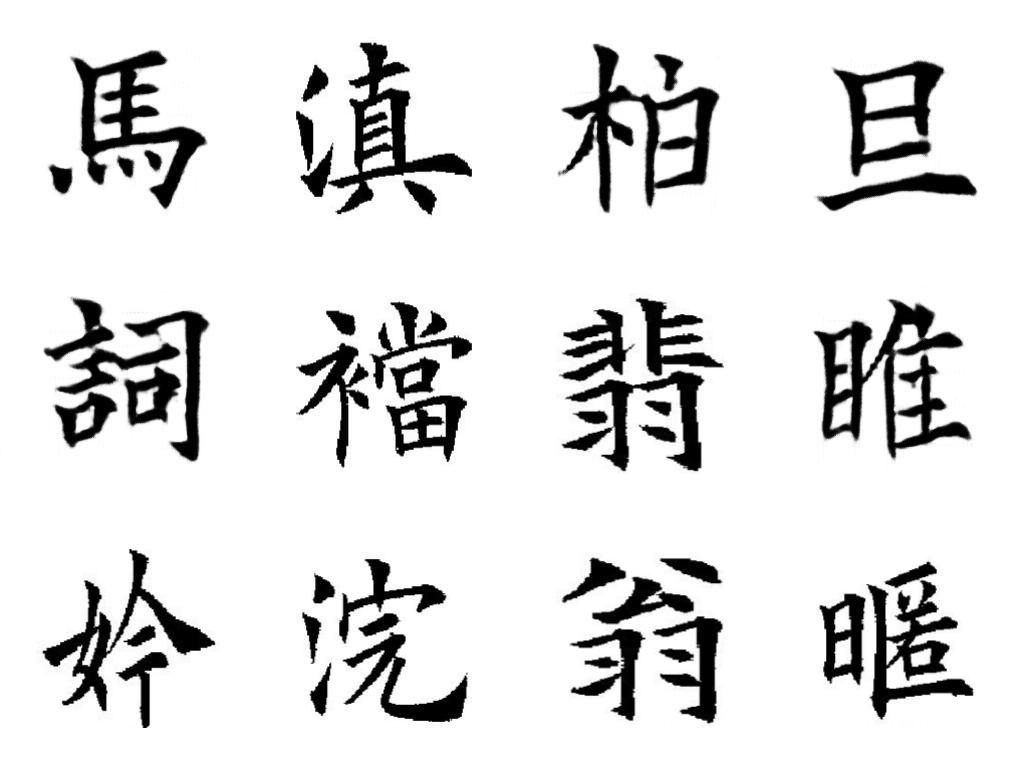}} &
        {\includegraphics[scale=0.093]{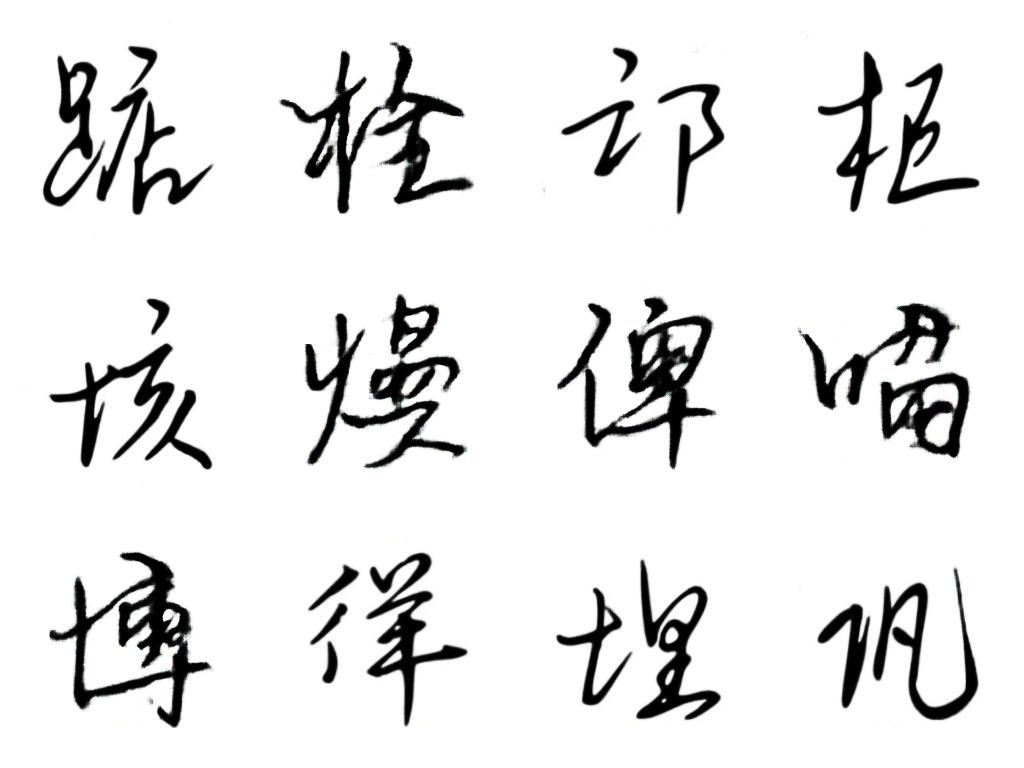}} &
        {\includegraphics[scale=0.093]{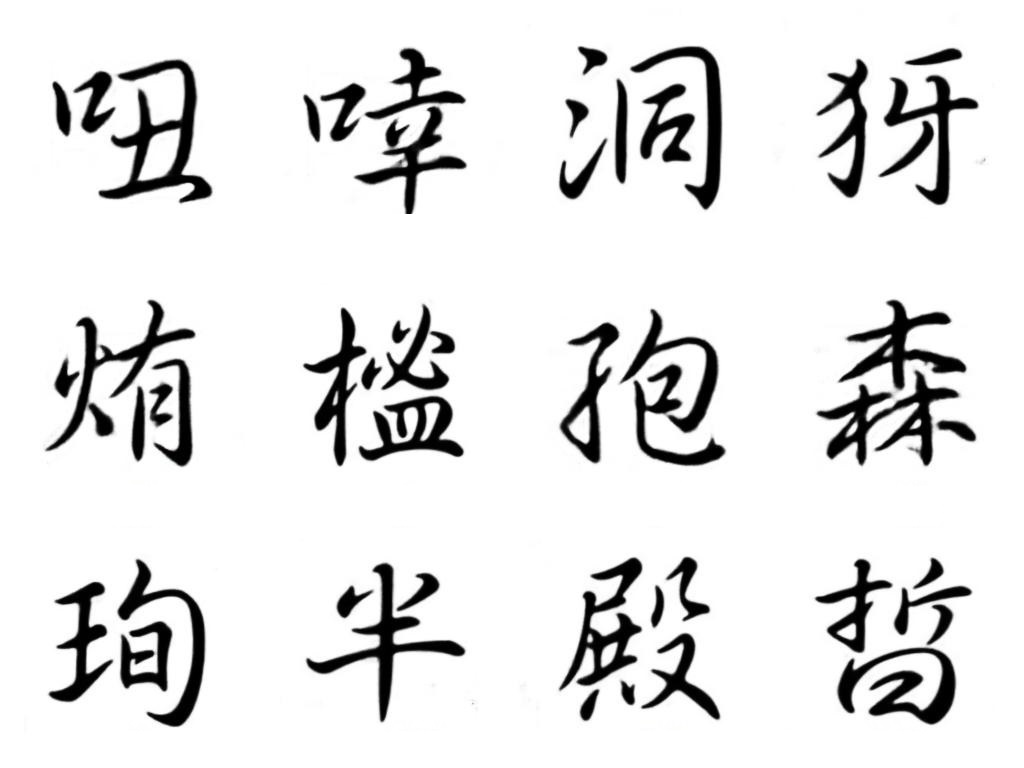}} \\
        \hline
    \end{tabular}}
\end{table}

\setlength{\textfloatsep}{5pt}
\subsection{Quantitative Evaluation}
For quantitative evaluation, we evaluate it from two aspects: style and content. For the former, we use Intersection Over Union(IOU) to measure whether our results have completed the style translation. IOU calculates the ratio of the intersection and union between the generated character images and the real character images. The higher the value indicates that the distribution of the generated images is closer to the distribution of the real images, and the result is better. As above, we compare six classic methods, and Table \ref{tab:IOU} shows that ZiGAN achieves the highest IOU scores.

Similarly, for content evaluation, we train a Resnet18 model as a Chinese character recognizer on the ground truth of all styles. And test it on the images generated by our test set. As we can see from Table \ref{tab:accuracy}, the top-1 accuracy achieved by our method is significantly ahead of other methods, which proves that our method can effectively retain the structure and content information of the character.

\setlength{\textfloatsep}{3pt}
\subsection{User Study}
We implement user study to verify that our results are not only better in the calculated indicators. 40 people who are familiar with Chinese characters participate in the experiment. We randomly select 65 characters in the test set, then use the compared methods and the proposed method to generate images. Therefore, the participants see a total of 520 images, including 390 images generated by the compared methods, 65 images generated by our ZiGAN, and 65 of ground truth. At each selection, participants will see 7 images generated by 7 different methods and ground truth. Overall, the participant’s goal is to find the image that is most similar to ground truth. In detail, participants are asked to prioritize finding the images that are semantically consistent with the ground truth, which means that the generated characters cannot have wrong radicals or missing strokes. On this basis, consider which image style is closer to the ground truth and has better details. Table \ref{tab:User study} shows the respondents' vote rates for each method.

Meanwhile, we build the Turing test set and make a Turing test. As shown in Table \ref{tab:Turing}, each sample contains 6 fake glyph images generated by ZiGAN and 6 real glyph images. We ask 50 professional Chinese users to identify which images are generated in 30 sets of samples. ZiGAN achieves an accuracy of 51.6\%, which is very close to random selection.

\begin{figure}[]
\centering
\includegraphics[scale=0.35]{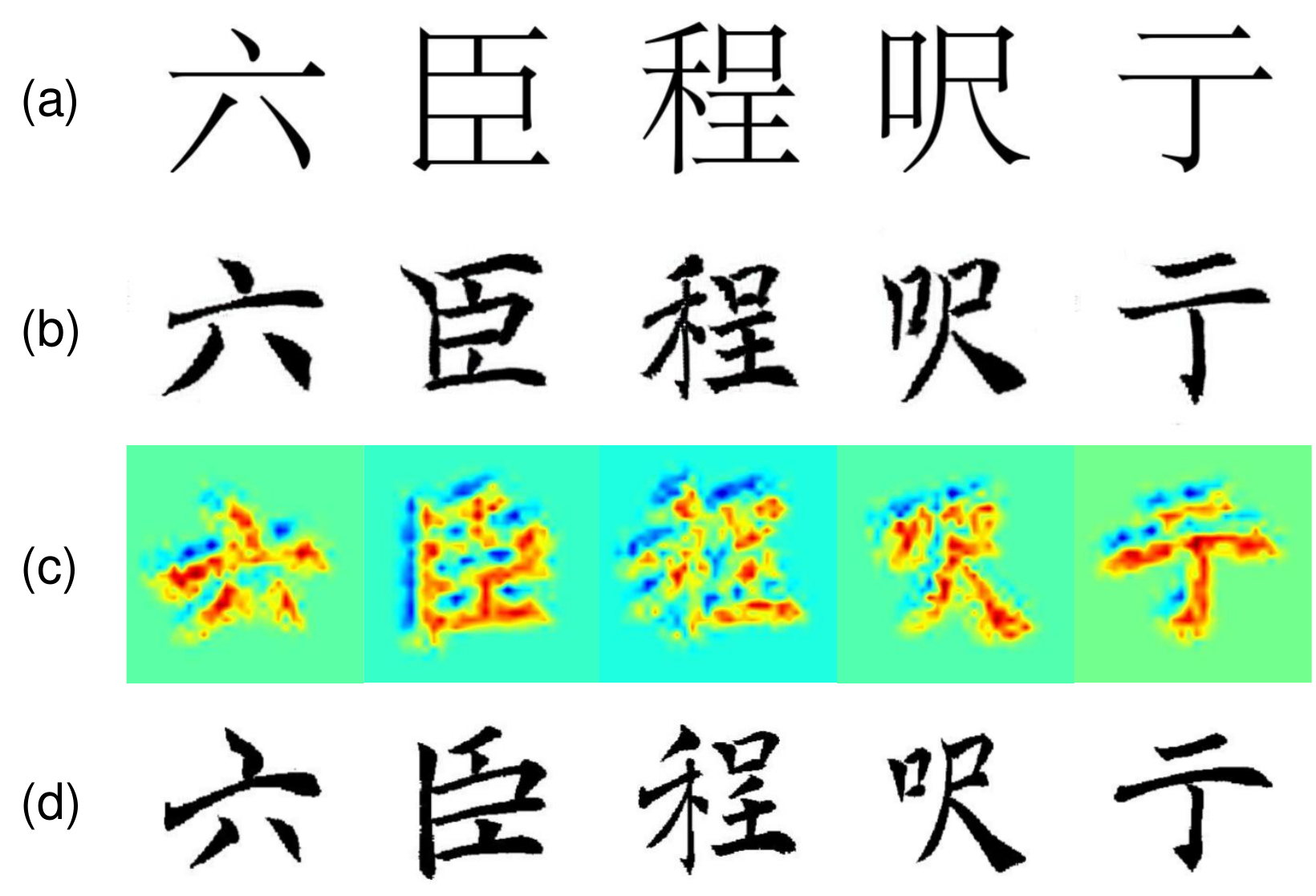}
\caption{Visualization of the attention maps:(a) Source style characters,(b) Generated target style characters,(c) Attention map of discriminator from source to target character,(d) The ground truth of target style characters.}
\label{fig:heatmap}
\end{figure}

\begin{figure}[]
\centering
\includegraphics[scale=0.27]{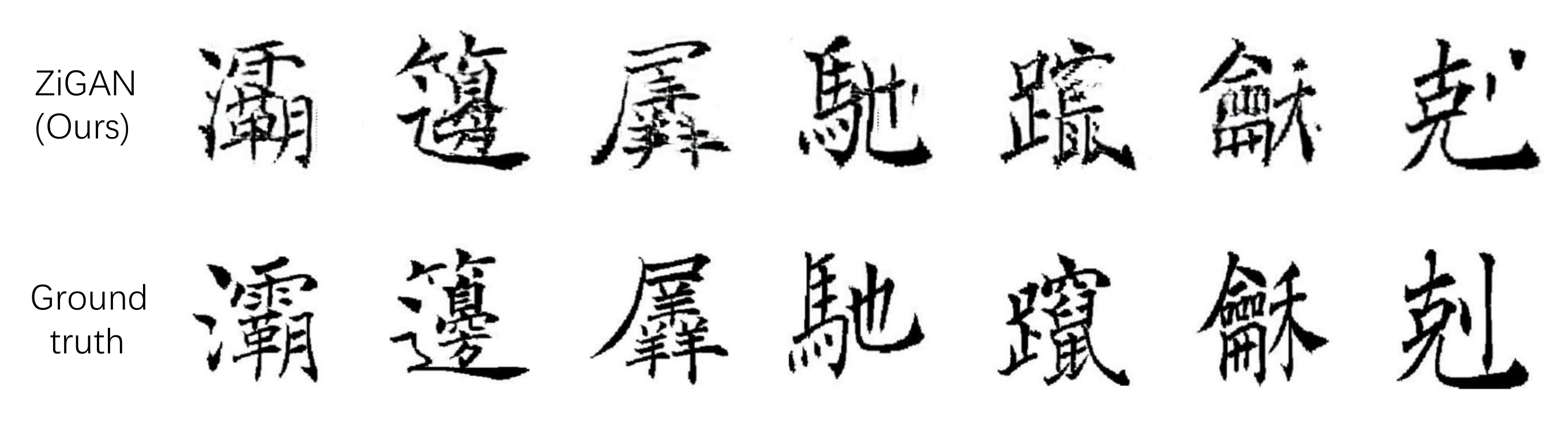}
\caption{Some unsatisfactory synthesis results.}
\label{fig:failure}
\end{figure}

\setlength{\textfloatsep}{5pt}
\subsection{Empirical Analysis}
\noindent\textbf{Ablation Studies}
In order to verify that each step in our framework is beneficial, we did an ablation experiment. The cam loss helps the discriminator to better distinguish the differences in the details of different character styles, while the generator can make improvements in the most important regions. The style loss helps our model learn additional style and structural knowledge of unpaired data, and innovatively align the distribution of unpaired source and target data in the feature space. The alignment loss maintains the generated image with intact semantic information from another level. The combination of these three forms our proposed method. Figure \ref{fig:xiaorong} displays the 200-shot image generation results without cam loss, style loss or alignment loss. Table \ref{tab:IOU} and Table \ref{tab:accuracy} also present the complete quantitative results of the ablation experiment. These results show that every module of our method is critical.

\noindent\textbf{Analysis of CAM Attention} We visualize the local attention maps of the discriminator in Figure \ref{fig:heatmap}. It shows which regions the discriminator focuses its attention to determine whether the target image is real or generated. In row(c) of Figure \ref{fig:heatmap}, we can find that this attention module has successfully found the main body of the characters, and pay more attention to the sharp strokes and radicals with high recognition. This is consistent with our intuition, people also distinguish font styles in this way.

\noindent\textbf{Failure cases} As shown in Figure \ref{fig:failure}, for some extremely complex characters, there are still some subtle deficiencies in the generated results. The lack of training data leads to poor generalization performance in complex situations. For future work, We are planning to work on using fewer target references and get more robust and generalized model.

\section{Conclusion}
In this paper, we propose a novel ZiGAN, which can accomplish fine-grained Chinese calligraphy font generation with few-shot references. The main idea is that extra structural knowledge can be learned by utilizing numerous unpaired characters. We also groundbreakingly align the feature distribution of different font styles to capture valuable style knowledge in target and strengthen the coarse-grained understanding of character content. Besides, our method is an end-to-end framework that does not require any manual operation or redundant preprocessing. It can be easily and quickly adapted to new tasks.


\clearpage

\balance
{
\bibliographystyle{ACM-Reference-Format}
\bibliography{{ZiGAN}}
}

\clearpage
\includepdfmerge{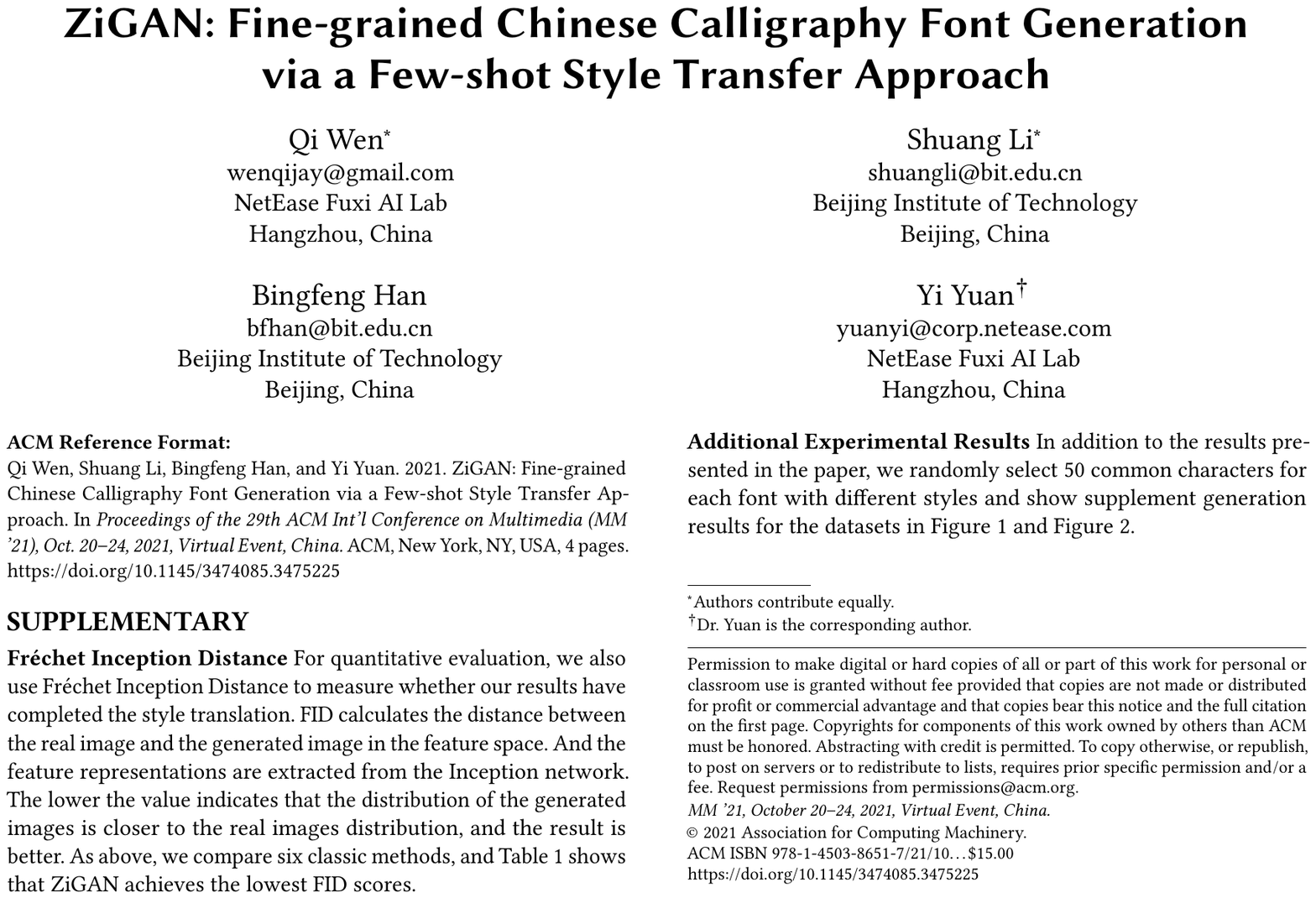,1-4}

\end{document}